\documentclass[conference]{IEEEtran}
\IEEEoverridecommandlockouts
\usepackage{cite}
\usepackage{amsmath,amssymb,amsfonts}
\usepackage{graphicx}
\usepackage{textcomp}
\usepackage{xcolor}
\usepackage{subfig}
\usepackage{tabularx}
\usepackage{ragged2e}
\usepackage{booktabs}
\usepackage{algorithmicx}
\usepackage{algpseudocode}
\usepackage{algorithm}

\def\BibTeX{{\rm B\kern-.05em{\sc i\kern-.025em b}\kern-.08em
    T\kern-.1667em\lower.7ex\hbox{E}\kern-.125emX}}
\begin{document}

\title{An End-to-End System for Culturally-Attuned Driving Feedback using a Dual-Component NLG Engine\\
\thanks{This work was supported by the Tertiary Education Trust Fund, Nigeria.}
}

\author{\IEEEauthorblockN{1\textsuperscript{st} Iniakpokeikiye Peter Thompson*}
\IEEEauthorblockA{\textit{Dept. of Computing Science} \\
\textit{University of Aberdeen}\\
Aberdeen, Scotland, United Kingdom \\
i.thompson.21@abdn.ac.uk}
\and
\IEEEauthorblockN{2\textsuperscript{nd} Dewei Yi}
\IEEEauthorblockA{\textit{Dept. of Computing Science} \\
\textit{University of Aberdeen}\\
Aberdeen, Scotland, United Kingdom \\
dewei.yi@abdn.ac.uk}
\and
\IEEEauthorblockN{3\textsuperscript{rd} Ehud Reiter}
\IEEEauthorblockA{\textit{Dept. of Computing Science} \\
\textit{University of Aberdeen}\\
Aberdeen, Scotland, United Kingdom \\
e.reiter@abdn.ac.uk}
}

\maketitle

\begin{abstract}
This paper presents an end-to-end mobile system that delivers culturally-attuned safe driving feedback to drivers in Nigeria, a low-resource environment with significant infrastructural challenges. The core of the system is a novel dual-component Natural Language Generation (NLG) engine that provides both legally-grounded safety tips and persuasive, theory-driven behavioural reports. We describe the complete system architecture, including an automatic trip detection service, on-device behaviour analysis, and a sophisticated NLG pipeline that leverages a two-step reflection process to ensure high-quality feedback. The system also integrates a specialized machine learning model for detecting alcohol-influenced driving, a key local safety issue. The architecture is engineered for robustness against intermittent connectivity and noisy sensor data. A pilot deployment with 90 drivers demonstrates the viability of our approach, and initial results on detected unsafe behaviours are presented. This work provides a framework for applying data-to-text and AI systems to achieve social good.
\end{abstract}


\section{Introduction}
Road traffic crashes are a critical public health crisis in low and middle-income countries (LMICs) \cite{who2023}. In Nigeria, this results in over 5,000 fatalities annually \cite{okpale2025}, driven by risky behaviours such as speeding and drink-driving \cite{labbo2024}. While technological interventions have shown promise in high-income countries \cite{braun2018,chen2023data}, they cannot be directly transferred due to challenges unique to the Nigerian context, including poor road infrastructure, unreliable connectivity, and a lack of digital road data \cite{uzondu2019,freedomhouse2021}.

This creates a significant engineering challenge: to build a system that is not only intelligent but also robust, context-aware, and culturally sensitive. This paper details such a system, focusing on its architecture and, most critically, its Natural Language Generation (NLG) engine designed to provide personalized feedback. 

The primary contribution of this work is the design and implementation of a dual-component NLG engine, which represents a unique contribution to the field. To the best of our knowledge, this is the first system to combine: (1) a legally-constrained, retrieval-augmented pipeline for generating factual safety 'Tips', with (2) a psychologically-grounded, two-step reflective pipeline for generating persuasive behavioural 'Reports'. This dual approach, specifically engineered for cultural-attunement and on-device deployment in a low-resource setting like Nigeria, presents a new framework for applied NLG.

\section{Related Work}
Our work is situated at the intersection of risky driving behaviour analysis and data-to-text generation. Research in driving analysis often focuses on identifying unsafe events from vehicle trajectory data \cite{zheng2022risky, mantouka2021}. While effective, such systems often assume high-quality data. In the Nigerian context, these approaches must be adapted to handle noisy sensor data and a lack of reliable ground truth \cite{olapoju2016}.

On the NLG side, data-to-text systems, a core area of Natural Language Generation \cite{nlg_survey_2018}, have become increasingly sophisticated. However, much research focuses on generating diverse outputs from structured data, often using constrained generation and ranking techniques \cite{chen2023data,garbacea_why_2025}. Our work extends these by focusing on a real-world application where the generated text must be not only be factually correct but also psychologically persuasive and culturally appropriate, a key challenge in creating effective behaviour change interventions \cite{orji2017persuasive}. We integrate a well-known and empirically validated psychological model, the Theory of Planned Behaviour (TPB) \cite{ajzen1991, armitage_efficacy_2001}, into our NLG pipeline. This theory, which posits that behaviour is driven by attitudes, norms, and perceived control, provides a strong foundation for crafting feedback that encourages genuine behaviour change, as it forms a core component of many integrated behavioural models \cite{braun2018, hagger_integrated_2014}. The challenge of adapting persuasive technologies to specific cultural contexts is a growing area of research \cite{orji2017persuasive}, and our work provides a concrete example of such an adaptation.

\begin{figure}[htbp]
\centering
\includegraphics[width=\columnwidth]{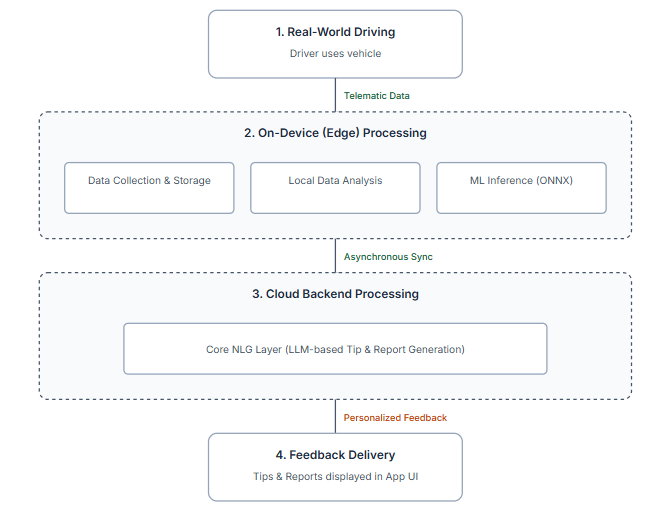}
\caption{Overall System Architecture and Application Flow.}
\label{fig:arch}
\end{figure}

\begin{figure}[htbp]
\centering
\includegraphics[width=\columnwidth]{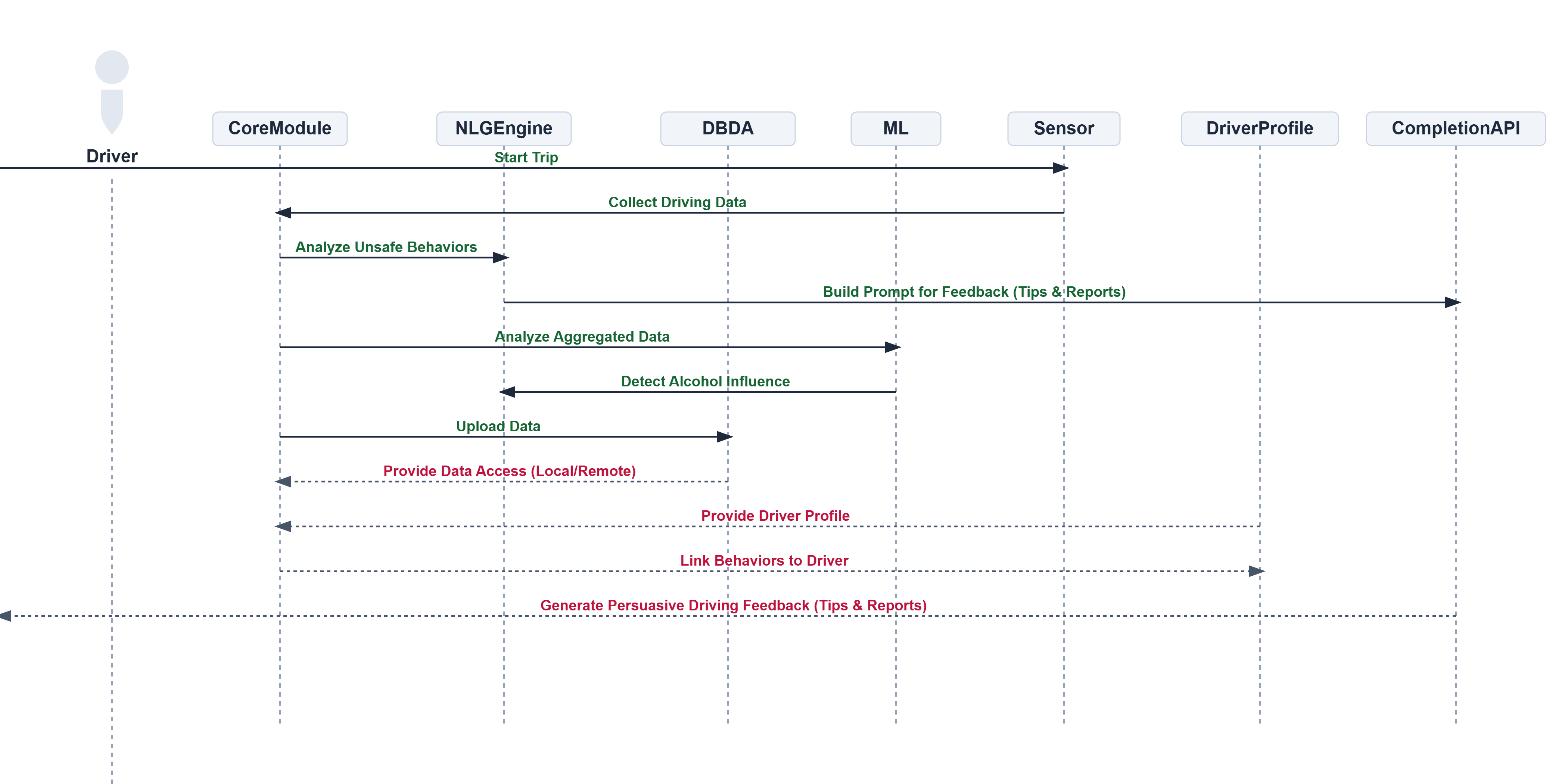}
\caption{Sequence diagram of component interactions.}
\label{fig:sequence_diagram}
\end{figure}

\section{System Architecture and Workflow}
Our system is an Android application built with a modern MVVM architecture. The workflow is designed to be fully automated, from detecting a trip's start to delivering final feedback.

\subsection{End-to-End Application Workflow}
The overall application flow is visualized in Fig.~\ref{fig:arch}. The process is managed by a foreground service that automatically detects when a trip starts and stops using a debouncing mechanism. Once a trip begins, a data collection service logs sensor data to a local Room database. At the end of the trip, a series of processing steps are initiated, as detailed in the sequence diagram in Fig.~\ref{fig:sequence_diagram}. This local-first architecture ensures the app is fully functional offline, preventing data loss and providing a seamless user experience in a low-connectivity environment.

\subsection{Behaviour Detection Components}
\subsubsection{Rule-Based Unsafe Event Detection}
In addition to the ML model, the system employs a real-time, rule-based engine to detect common unsafe events by monitoring established safety indicators \cite{niezgoda_measuring_2012}. Sensor data is continuously monitored against dynamic thresholds. These thresholds are not static; for instance, the sensitivity for detecting harsh braking is increased at higher speeds, adapting to the driving context. If a threshold is crossed, a corresponding unsafe behaviour event is created and stored locally.

\subsubsection{Machine Learning Component for Alcohol Detection}
To address the specific local issue of drink-driving, a specialized classification model was developed \cite{thompson_mobile_nodate}. As this model's development is detailed elsewhere, we provide only a summary here. We used a Decision Tree classifier, which provides a high degree of interpretability required for this safety-critical application. Trained on a real-world dataset collected in Nigeria and balanced using SMOTE, the model's logic relies on five key features derived from driving patterns (mean hour of trip, day of the week, standard deviation of speed and course, and mean Y-axis acceleration). The final model achieved 100\% recall on the test set, correctly identifying all alcohol-influenced trips. For on-device deployment, the model was converted to the ONNX format and runs via the ONNX Runtime, with an average inference time under 50ms on a mid-range Android device.

\subsection{The Dual-Component NLG Engine}
\begin{figure}[htbp]
\centering
\includegraphics[width=\columnwidth]{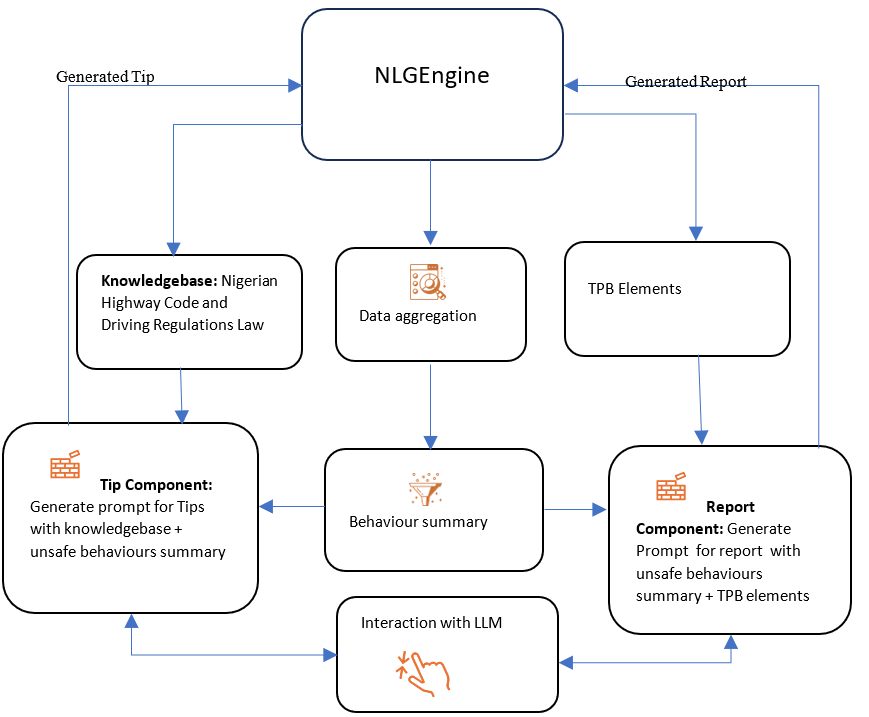}
\caption{High-level architecture of the dual-component NLG engine, showing the parallel pipelines for Tips and Reports.}
\label{fig:nlg_components}
\end{figure}

\begin{figure}[htbp]
\centering
\includegraphics[width=0.55\columnwidth]{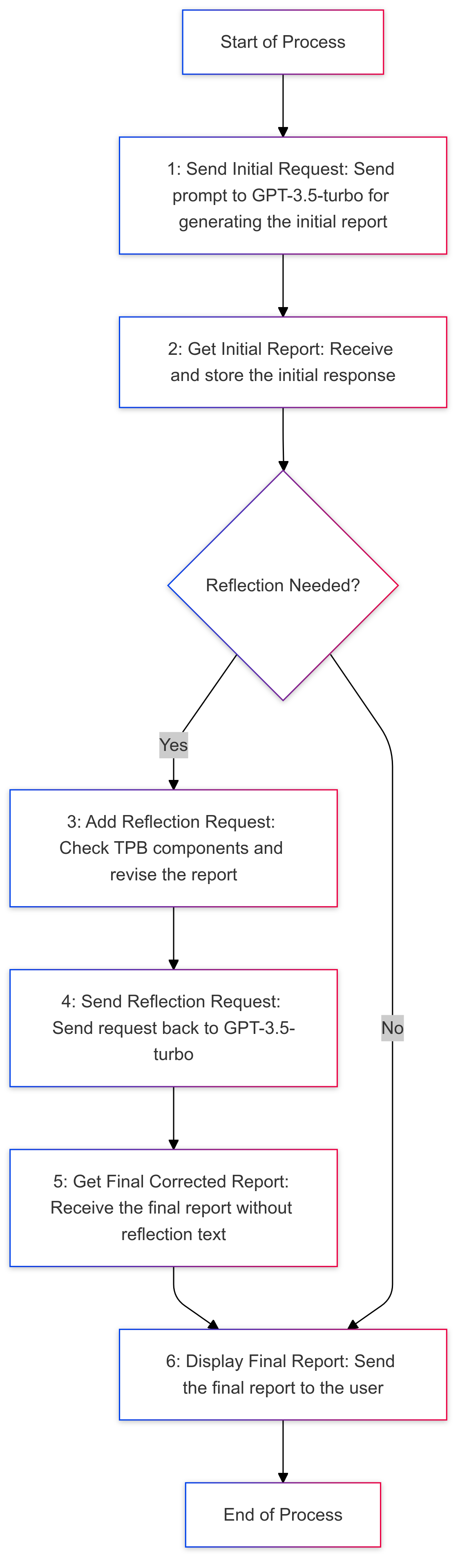}
\caption{The two-step reflection process for generating persuasive reports.}
\label{fig:report_process}
\end{figure}

The core of our system is an NLG engine composed of two distinct pipelines for generating different types of feedback. The overall architecture is visualized in Fig.~\ref{fig:nlg_components}, while the process for generating reports is detailed in Fig.~\ref{fig:report_process}.

\subsubsection{The NRAG 'Tips' Component}
To provide legally-grounded advice, a Naive Retrieval-Augmented Generation (NRAG) pipeline is implemented. We chose a 'naive' keyword-based retrieval method over more complex semantic search as an intentional engineering trade-off. While semantic search offers flexibility, it carries a risk of retrieving a regulation that is contextually similar but legally incorrect, which is unacceptable in a safety application. Our keyword-based approach guarantees 100\% factual grounding and explainability by mapping a detected behaviour directly to a pre-defined section of a local JSON file containing the Nigerian Highway Code. This factual context is then injected into a prompt for an LLM, which is instructed to generate a concise safety tip. This method is a practical solution to the open challenge of constrained language generation, where the primary goal is to enforce hard, testable constraints on the output text to ensure its factual accuracy \cite{garbacea_why_2025}.

\subsubsection{The Reflective 'Reports' Component}
To provide holistic, persuasive feedback, the system generates weekly reports via a sophisticated \textit{two-step reflection process} (Fig.~\ref{fig:report_process}). This process was designed to mitigate known failure modes of LLMs, namely factual inconsistency. The selection of models was based on a trade-off between performance and cost.
\begin{enumerate}
    \item \textbf{Initial Draft Generation:} A detailed prompt with driver statistics and TPB-grounded instructions is sent to `gpt-4-turbo`, chosen for its strong reasoning and high-quality text generation capabilities, to produce a comprehensive initial draft.
    \item \textbf{Refinement and Correction:} The draft is then passed to `gpt-4o-mini`, a faster and more cost-effective model, with a specific revision instruction. This model is well-suited for the narrower task of refinement, ensuring the final report is factually consistent and adheres to the desired supportive tone.
\end{enumerate}
This hybrid approach is a practical solution for deploying LLMs in applications requiring high factual fidelity without incurring the cost of using the most powerful model for all stages.
\begin{figure*}[t]
\centering
\subfloat[An example of a generated legal tip.]{\includegraphics[height=7.5cm]{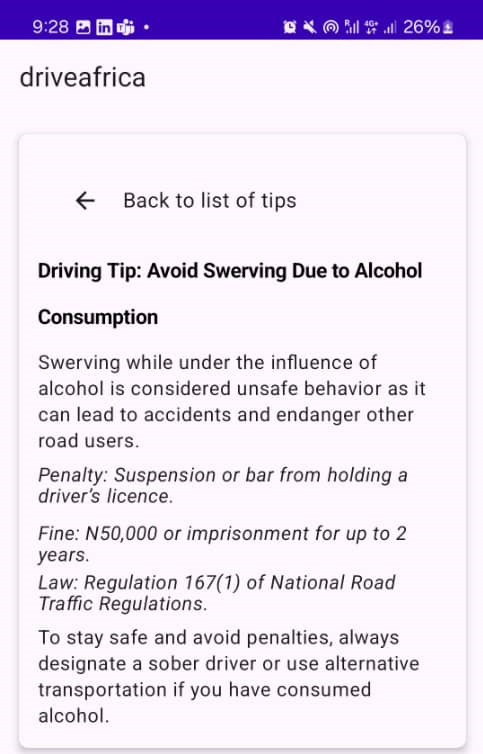}%
\label{fig:tip_example}}
\hfill
\subfloat[Part 1 of a generated weekly report.]{\includegraphics[height=7.5cm]{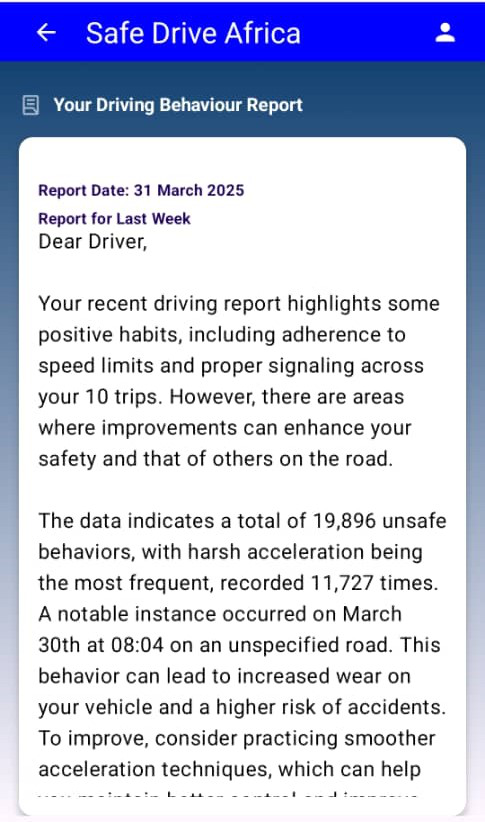}%
\label{fig:report1}}
\hfill
\subfloat[Part 2 of the same weekly report.]{\includegraphics[height=7.5cm]{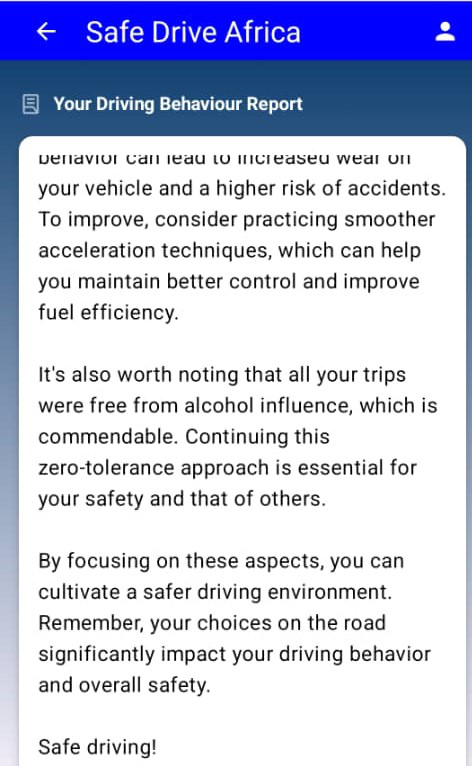}%
\label{fig:report2}}
\caption{Examples of final system outputs as displayed in the mobile application.}
\label{fig:output_examples}
\end{figure*}
\section{System Deployment and Initial Evaluation}
The system was deployed in a pilot study with 90 drivers across four Nigerian cities, collecting over 2,000,000 data points. The pilot served as a successful real-world test of the system's data pipeline and robustness.

\subsection{Initial Pilot Study Results}
While a full analysis of behaviour change is ongoing, the initial data provides valuable insights into driving patterns. The rule-based detection engine logged a total of 154,230 unsafe events during the pilot period. The distribution of the most common events is shown in Table~\ref{tab:unsafe_events}. Harsh acceleration, often associated with aggressive driving in congested traffic, was the most frequently detected event. Speeding was the second most common, highlighting a widespread disregard for posted limits. This initial analysis confirms that the system is effective at capturing the specific types of risky behaviours prevalent in this driving context.

\begin{table}[htbp]
\caption{Distribution of Detected Unsafe Events from Pilot Study}
\centering
\begin{tabular}{@{}lr@{}}
\toprule
\textbf{Unsafe Behaviour} & \textbf{Frequency (\%)} \\
\midrule
Harsh Acceleration & 46.1\% \\
Speeding & 29.5\% \\
Harsh Braking & 18.3\% \\
Swerving & 6.1\% \\
\bottomrule
\end{tabular}
\label{tab:unsafe_events}
\end{table}

The pilot deployment also surfaced practical engineering challenges, such as variability in smartphone sensor quality across devices and the need to manage battery consumption. Future iterations will focus on addressing these through calibration routines and service optimizations. When the full study period concludes, we will analyse the data for evidence of behaviour change.

\subsection{Qualitative Feedback Examples}
The primary output of the system is the generated text. Fig.~\ref{fig:output_examples} shows examples of the final feedback displayed to the user, including a legally-grounded tip and a persuasive weekly report, demonstrating the output of the dual-component NLG engine.

\section{Future Evaluation Plan}
To rigorously evaluate the system's impact on driver behaviour, a comprehensive mixed-methods study is planned for the conclusion of the current pilot deployment. This evaluation is designed to address the project's primary research questions regarding the effectiveness of the feedback in reducing unsafe driving practices.

\subsection{Study Design}
The evaluation will follow a \textbf{pre-post intervention study design}, comparing driver behaviour during a baseline period (where no feedback is provided) with an intervention period (where personalized feedback is enabled). The analysis will use data from the 90 drivers currently enrolled in the pilot study.

\subsection{Quantitative Analysis}
The primary goal of the quantitative analysis is to statistically measure changes in driving patterns using the sensor data collected by the application. The planned statistical tests include:
\begin{itemize}
    \item \textbf{Paired t-tests} or their non-parametric equivalent, the \textbf{Wilcoxon signed-rank test}, will be used to compare the mean frequency of unsafe events (e.g., harsh braking, speeding) per driver between the baseline and intervention phases.
    \item \textbf{Repeated Measures ANOVA} will be employed to analyze behavioural trends over multiple time points throughout the entire study period.
    \item \textbf{Correlation Analyses} will be used to compare the objective sensor data with self-reported behaviours gathered from driver surveys to check for consistency.
\end{itemize}

\subsection{Qualitative Analysis}
To understand the contextual factors influencing the effectiveness of the feedback, a qualitative analysis will be conducted. This will involve conducting \textbf{semi-structured interviews} with a subset of the participating drivers and analyzing the interview transcripts using \textbf{Thematic Analysis}. The goal is to identify key themes related to app usability, feedback clarity, user satisfaction, and overall persuasiveness.

This dual approach will provide a holistic understanding of not only \textit{if} the app changes behaviour, but also \textit{how} and \textit{why} it succeeds in the real-world context of Nigerian drivers.

\section{Discussion and Limitations}
The success of the domain and socio-cultural knowledge-driven approach provides a significant insight. Findings show that by incorporating qualitative data from driver surveys to engineer features, a more predictive model could be built \cite{thompson_mobile_nodate}. This underscores a key takeaway for data science in socio-technical domains: a hybrid approach that fuses human expertise with data-driven techniques can yield superior results.

Several limitations are acknowledged. The number of cities (four) for the pilot study was limited. Future work will focus on expanding the dataset for the ML model and the geographical scope of the evaluation.

\section{Conclusion}
This paper presented the detailed design and implementation of an end-to-end system that uses a novel dual-component NLG engine to provide culturally-attuned safe driving feedback in Nigeria. We detailed the system architecture, from automatic trip detection and on-device AI to a sophisticated, theory-driven feedback generation pipeline. The primary contribution is a complete mobile NLG-based feedback system engineering framework that demonstrates how to engineer a robust data-to-text application for a challenging, low-resource environment, providing a practical blueprint for applying advanced NLP and software engineering to solve critical safety problems.

\section*{Acknowledgment}

We would like to sincerely thank all the drivers who participated in the pilot testing of the app. We also appreciate our research collaborators including MaxExpress, Kabiru Momodu, and Abraham Desire, for helping to recruit the drivers involved in the experiment.

Special thanks go to the Tertiary Education Trust Fund (TETFund) for funding the larger PhD project that made this work possible.

\bibliographystyle{IEEEtran}
\bibliography{ref}

\end{document}